# Sāmayik: A Benchmark and Dataset for English-Sanskrit Translation

**Ayush Maheshwari**[1*], **Ashim Gupta**[2], **Amrith Krishna**[3], **Atul Kumar Singh**[4],
**Ganesh Ramakrishnan**[4], **G. Anil Kumar**[5], **Jitin Singla**[5]

[1]Vizzhy Inc Bengaluru, [2]University of Utah, [3]Learno.ai,
[4]Indian Institute of Technology Bombay, [5]Indian Institute of Technology Roorkee

**Abstract**

We release Sāmayik, a dataset of around 53,000 parallel English-Sanskrit sentences, written in contemporary prose. Sanskrit is a classical language still in sustenance and has a rich documented heritage. However, due to the limited availability of digitized content, it still remains a low-resource language. Existing Sanskrit corpora, whether monolingual or bilingual, have predominantly focused on poetry and offer limited coverage of contemporary written materials. Sāmayik is curated from a diverse range of domains, including language instruction material, textual teaching pedagogy, and online tutorials, among others. It stands out as a unique resource that specifically caters to the contemporary usage of Sanskrit, with a primary emphasis on prose writing. Translation models trained on our dataset demonstrate statistically significant improvements when translating out-of-domain contemporary corpora, outperforming models trained on older classical-era poetry datasets. Finally, we also release benchmark models by adapting four multilingual pre-trained models, three of them have not been previously exposed to Sanskrit for translating between English and Sanskrit while one of them is multi-lingual pre-trained translation model including English and Sanskrit. The dataset and source code is present at https://github.com/ayushbits/saamayik.

## 1. Introduction

We release Sāmayik, an English-Sanskrit parallel dataset, that covers the contemporary usage of Sanskrit, written in prose. Sāmayik is a Sanskrit term that translates to the "sayings of the contemporary world". Sāmayik consists of 52,961 parallel sentence pairs, collected from five different sources. These are spoken content that covers contemporary world affairs, interpretation of literary works, pedagogical content, *etc*.

'Itihāsa' currently forms the largest parallel machine translation corpus in English-Sanskrit (Aralikatte et al., 2021). This Sanskrit-English dataset comprises 93,000 pairs of verses in Sanskrit along with their corresponding English translations. These Sanskrit verses belong to Rāmāyaṇa and Mahābhārata written in the poetry form and belong to the classical era literature.

Sanskrit is estimated to have around 30 million extant manuscripts fit for digitization. Moreover, it has more than two million active speakers (McCartney, 2019; Chandramouli, 2011). Despite its rich heritage, Sanskrit remains classified as a low-resource language with no more than one million monolingual sentences available in the digitized form (Hellwig, 2010–2021; Maheshwari et al., 2022). The available digitized corpora for Sanskrit are vastly diverse not just in terms of the domains and chronology they span, but also in terms of the usage, stylistic features, the underlying syntax (Hellwig, 2009), and even the typological characteristics such as word order (Krishna et al., 2021; Tubb and Boose, 2007).

Sentence constructions in Sanskrit follow relatively free word order. Here, sentences written in verse form have to adhere to prescribed meter patterns as per prosody. Hence, word order need not adhere to a fixed word-order pattern. However, sentences written in prose tend to form Subject-Object-Verb (SOV) ordering. Ithihāsa and other monolingual available corpora predominantly represent content written in poetry form. Content written in prose is generally underrepresented in available digitized corpora in Sanskrit, especially those written in the contemporary era. To bridge this gap we release Sāmayik.

Sāmayik is a parallel Sanskrit-English dataset encompassing multiple contemporary corpora, providing a comprehensive representation of the contemporary usage of Sanskrit. In Section 2, we provide a detailed description of each source included in our dataset, and Table 1 presents an overview of the statistics for each source. The latest corpus in our collection contains content as recent as 2022, from an ongoing podcast series 'Mann Ki Baat'. The oldest corpus in our collection is the English-Sanskrit Bible, where the Sanskrit translation was performed in 1851 and it forms less than 14% of the overall dataset. The Sanskrit component in the rest of the corpora is composed either in the latter half of the twentieth century or in the current century.

In addition to our dataset, we release three benchmarks by adapting pre-trained models for neural machine translation in English-Sanskrit and vice-versa. Here, we adapt four pre-trained multi-lingual seq2seq models for the task, namely ByT5 (Xue et al., 2022), mBART (Liu et al., 2020), IndicBART (Dabre et al., 2022), and Indictrans (Gala et al., 2023). Except Indictrans, none of the models were exposed to Sanskrit during their pretraining stage where Indictrans is a multi-lingual translation model including English and Sanskrit.

---

*Outcome of research while pursuing PhD at IIT Bombay.
Correspondence: {ayusham, ganesh}@cse.iitb.ac.in

## 2. Sāmayik

Sāmayik is an English-Sanskrit machine translation dataset, consisting of 52,941 sentences from five different corpora. The aim of the dataset is to include translation pairs containing Sanskrit prose written in the modern era. We hired several professional English and Sanskrit linguistic experts for the purpose of translation and alignment for the development of the dataset. The educational qualifications of the experts range from a Master's degree to a Ph.D. The experience of the experts ranged from 3-20 years, with the more experienced ones assigned the job of translation while junior members were assigned the job of sentence alignment. The experts were paid as per the norms laid out by the norms set by the Government of India. Below, we give a brief description of each of the datasets involved and the steps involved in processing these sentences.

1. **Bible - The New Testament:** We release the New Testament of the Bible aligned with its corresponding English version. We use the Sanskrit version released by Calcutta Baptist Missionaries, originally published in 1851[1]. The New Testament contains 7,838 sentences from 260 chapters. Each verse is generally indexed by the book name, and chapter name followed by the verse number. For the English version of the Bible, we rely on Christodouloupoulos and Steedman (2015) where the English sentences also follow the same indexing form. Given the one-to-one correspondences at the sentence level for both English and Sanskrit sentences, the mapping was straightforward. We finally obtained a total of 7,838 parallel sentences. Further, three fluent speakers of both English and Sanskrit have verified the alignments for of 100 sentences, randomly sampled from the corpus.

2. **Mann ki Baat (MKB)**[2] - MKB is an ongoing monthly radio podcast hosted by the Prime Minister of India, which resumed its broadcast in 2014. Each episode is an address to the nation discussing social, cultural, and contemporary topics including conversation with individuals. Sanskrit translations by experts, albeit unofficial, are available in public domain[3]. We use these expert translations and manually align Sanskrit sentences with official English transcripts from 25 episodes. Additionally, these Sanskrit translations are further verified by 3 in-house language experts. The MKB English-Sanskrit corpus Sāmayik release consists of 4,047 sentences with a total of 47,843 words.

3. **Gītā Sopānaṁ** - Gītā Sopānaṁ is a book published by 'Samskrita Bharati' in 2009 for teaching Sanskrit to beginners. It consists of a total of 6130 sentences. As observable in Table 1, the count of unique words is just 6465 for these 6130 sentences. Gītā Sopānaṁ is a self-learning book targeted at beginners and enables them to learn Sanskrit through stories. It often contains simple and small sentences with a focus on learning the grammar instead of expanding vocabulary. We perform in-house translation of the work to English sentences with the help of 4 language experts well-versed in both English and Sanskrit. Given the expert-level annotations, we only gather one translation per Sanskrit sentence. In summary, each expert annotated around 1500 sentences.

4. **Spoken Tutorials**[4] - Spoken Tutorial project is a large corpus of video tutorials for training students to use open-source software. These tutorials are created by domain experts and translated into several languages by expert translators. We scraped[5] videos and transcripts from their website for which both English and the corresponding Sanskrit translations are available. We extracted transcripts of 254 videos where each video is an of average 10 minutes in duration. The transcripts are manually created and, therefore, do not require additional sentence segmentation. The alignment between the English and the corresponding sentences for each transcript was performed manually with the help of 5 linguistic experts. We ask experts to align English and Sanskrit sentences from the transcripts and merge sentences if one-to-one correspondence is not present. Each expert aligned around 5,000 sentences. The final corpus contains 23,835 sentences comprising 237,449 words.

5. **NIOS** - The National Institute of Open Schooling (NIOS) is a national-level board of education in India established in 1989. NIOS prints self-instructional study materials for various subjects up to the senior secondary education level. We obtained the study materials from the Indian knowledge tradition courses offered by NIOS, which are available in both English and Sanskrit[6]. Each course consists of multiple topics accessible in the form of PDF files. We use PDF parsers to convert PDF content in text format, without loss of information. We hired a team of five English and Sanskrit linguistic experts who aligned the sentences from the corresponding text files. NIOS contains 11,356 parallel sentences with 105,178 total words and 30,966 unique words.

---

[1] https://www.bible.com/bible/2104/MAT.1.SAN-DN
[2] https://pmonradio.nic.in/
[3] https://sanskritdocuments.org/sites/manogatam/
[4] https://spoken-tutorial.org/
[5] The website content is licensed under CC4.0 license.

| Dataset | NIOS | Spoken Tutorials | GitaSopanam | Bible | Mann Ki Baat | Total |
| --- | --- | --- | --- | --- | --- | --- |
| #sentences | 11356 | 23835 | 5885 | 7838 | 4047 | 52941 |
| #words | 105178 | 237449 | 26135 | 102508 | 47843 | 518842 |
| #unique words | 30966 | 38373 | 6513 | 38359 | 20484 | 122349 |
| % of unique words | 29.4 | 16.2 | 24.9 | 37.4 | 42.8 | 23.6 |
| Mean word length | 9.3 | 10 | 4.5 | 13.1 | 11.8 | 9.8 |

Table 1: Number of sentences, words, unique words and average word length for different corpus in Sāmayik.

## 3. Experiments

### 3.1. Systems

1. **mBART** (Liu et al., 2020): is a multilingual pretrained seq2seq model trained using similar objective as employed in BART (Lewis et al., 2020). We employ mbart-large-50-many-to-many-mmt, trained on a large multilingual corpus of 50 languages, for our experiments. The vocabulary size of the pre-trained model is 250K and maximum sequence length of 1024 with 610M parameters.

2. **IndicBART** (Dabre et al., 2022) is a multilingual pretrained seq2seq model with 244M parameters trained using the pre-training objective of BART. IndicBART was trained using corpora from Indic languages and English. While different Indic languages use different scripts, these are losslessly converted to Devanagari before tokenization during its pretraining. Hence, we use the Devanagari script for encoding Sanskrit, and Roman script for English.

3. **ByT5** (Xue et al., 2022) is a token free pretrained seq2seq model following the pre-training objective of mT5 (Xue et al., 2021). However, here it is a token-free model that uses a fixed 256-byte value in Unicode as its vocabulary. From prior work (Maheshwari et al., 2022), we observe that the use of the Devanagari script in Unicode to encode content in Sanskrit leads to the best results. We use a base version of ByT5 in our experiments which consists of 582M parameters where UTF-8 bytes are directly fed into the model without any text pre-processing.

4. **IndicTrans** (Gala et al., 2023) is a multi-lingual translation model trained on 22 Indic languages including Sanskrit. The multi-lingual model is trained with the English-Sanskrit bi-text pairs. The NLLB corpora of 3M sentences were filtered to remove noisy sentence pairs. The sentence pairs were filtered using margin-based scoring that finds the closest semantic match between the pairs of source and target sentences. Finally, the model is trained with a dataset size of 244,367 English-Sanskrit bi-lingual sentence pairs. The model is trained with the transformer architecture comprising of 18 encoder and 18 decoder layers with the feedforward dimension of 8192, and 16 attention heads. The model uses sub-word tokenization with the maximum vocab size of 32K for English-Sanskrit and 128K for Sanskrit-English models and parameter count of 1.1B. We fine-tune the English-Sanskrit and Sanskrit-Eng model with Sāmayik corpus.

### 3.2. Experimental Setup

**Metrics**: We evaluate the performance of the models on both BLEU (Papineni et al., 2002) and ChrF (Popović, 2015). BLEU is a word-level n-gram precision-based metric whereas ChrF is a character-level n-gram F-score. Here, given that Sanskrit is a morphologically rich language with more than 1,400 possible inflected forms (Krishna et al., 2021), we believe ChrF can be indicative of capturing morpho-syntactic aspects.

**Data**: Due to the relatively low availability of the data, we consider 90% of sentence pairs from the four corpora NIOS, Spoken Tutorial (ST), Gita Sopanam (GS), and Bible as our training set and the rest as our in-domain evaluation set. The evaluation set is equally split into development and test set. To evaluate the performance of our model on a completely different domain test set, we reserve Mann Ki Baat(MKB) as an out-of-domain test set, implying that MKB was not included in the training data.

**Implementation Details**: All models are fine-tuned from their pre-trained checkpoints using HuggingFace Transformers (Wolf et al., 2020). Both source and target sequences are truncated at 512 token lengths and set the batch size to 128. We use the standard cross entropy loss with label smoothing of 0.1 and AdamW optimizer (Loshchilov and Hutter). All model are trained for a maximum of 30 epochs with a batch size of 16, learning rate of 1e-3, label smoothing factor of 0.1 and weight decay of 1e-4. For IndicTrans, the learning rate is set to 1e-4, dropout is 0.2 and maximum tokens per batch of 1024 and patience of early stopping was set to 5.

---
[6] https://www.nios.ac.in/online-course-material/indian-knowledge-tradition.aspx

|        | En-Sa |       | Sa-En |       |
|--------|-------|-------|-------|-------|
| **Model** | **BLEU** | **ChrF** | **BLEU** | **ChrF** |
| ByT5 | **28.7** | 44.4 | 31.1 | 55.7 |
| mBART | 27.20 | **46.61** | 11.6 | 27.0 |
| IndicBART | 25.45 | 43.47 | 29.79 | 50.14 |
| Indictrans | 11.3 | 46.6 | **37** | **58.2** |

Table 2: Results for different models on the in-domain test set for En-Sa and Sa-En direction.

|        | En-Sa |       | Sa-En |       |
|--------|-------|-------|-------|-------|
| **Model** | **BLEU** | **ChrF** | **BLEU** | **ChrF** |
| ByT5 | 7 | 21.4 | 5.4 | 29 |
| mBART | **7.11** | 22.6 | - | - |
| IndicBART | 6.9 | 22.4 | 5.3 | 27.7 |
| Indictrans | 0.6 | **26.7** | 13.1 | 37.5 |
| Google Trans | 1.9 | 35 | **13.9** | **44.7** |
| NLLB | 1.2 | 27.6 | 11.5 | 36.1 |
| Indictrans(Vanilla) | 1.2 | 34 | 14.5 | 42.7 |

Table 3: Results for out-of-domain test set, namely, Mann Ki Baat (MKB) for En-Sa and Sa-En directions. We omit mBART due to poor performance on in-domain test split (refer Table 2). Performance reported on the Google Translate, NLLB and Indictrans (Vanilla) (below double horizontal line) refers to the evaluation with pre-trained models.

### 3.3. Results

Table 2 shows the performance of all four systems on the in-domain test set. These systems are fine-tuned on the in-domain training data. As it can be observed from the table, different models perform the best depending on the direction of the translation. Here, mBART reports the best results for English-Sanskrit (En-Sa) translation, whereas Indictrans performs the best for Sanskrit-English (Sa-En) translation. Despite being pre-trained on significant amount of English-Sanskrit parallel corpus, Indictrans reports lower scores for En-Sa direction. However, model reports better scores on Sa-En direction. We hypothesise this can be attributed to the high morphological characteristics of the Sanskrit language which prevents fair evaluation using existing metrics. ByT5 reports the second best results for En-Sa, though on an average it requires more than five times the sequence length than that of the other models. Here, ByT5 reports a sequence length of 156.99, as against 30 for the model with the next longest sequence length, mBART. The disparity in sequence length arises out of the tokenizers used in ByT5 which is at a Unicode byte level against the subword tokenizers used in the other models.

| Model | IndicBART | | mBART | |
|-------|-----------|---|-------|---|
| Dataset | BLEU | ChrF | BLEU | ChrF |
| Itihasa | 4.6 | 16 | 4.3 | 16.7 |
| Sāmayik | 6.3 | 23.2 | **7.3** | **22.3** |
| Itihasa + Sāmayik | **6.9** | **22.4** | 6.8 | 21.6 |

Table 4: Comparison between existing dataset Itihasa, Sāmayik and Itihasa + Sāmayik on MKB out-of-domain testset for English-Sanskrit translation direction. The score difference between Itihasa (1st row) and Sāmayik(2nd row) are statistically significant at p<0.05.

We perform a zero-shot evaluation using MKB, on our out-of-domain test data, not just on the four systems fine-tuned on Sāmayik, but also on three publicly available systems, namely Google translate (GT) service[7], NLLB-200 1.3B variant[8], and IndicTrans with no fine-tuning. As shown in Table 3, GT outperforms all other systems by a considerable margin in the Sa-En direction, though a significant drop is observed in the En-Sa direction. However, for the En-Sa direction, our fine-tuned version of mBART performs the best in terms of BLEU and our fine-tuned version of Indictrans in terms of ChrF.

The considerable drop in performance on the in-domain dataset may be attributed to the vocabulary diversity generally observed in Sanskrit corpora. Sanskrit corpora tend to have a long tail of rare words within the corpus. Further, owing to high lexical productivity both with compounding and derivation, these corpora tend to have a diverse vocabulary for one another. Both the long tail of rare words and rich compounding are challenging for models, similar to the current NMT models, that rely on distributional semantics.

'Contemporyness' forms a key factor for the corpora in Sāmayik. Table 4 shows the performance on MKB in En-Sa translation on the mBART and IndicBART, fine-tuned using an alternate publicly available dataset 'Itihasa'. Here, Itihasa has nearly double the number of training instances than Sāmayik. In spite of it, models fine-tuned on Sā-mayik outperform that on Itihasa for MKB. Further, model trained with the combination of Itihasa and Sāmayik report marginally better scores on En-Sa than the models trained only using SāmayikOn the contrary in Sa-En, Sāmayik reports better results than the combination of two datasets. We find that systems trained using our dataset has significantly higher BLEU and ChrF scores reinforcing the need for a corpus that follows contemporary content in Sanskrit.

---
[7]Accessed via https://translation.googleapis.com/language/translate/v2
[8]Accessed via https://huggingface.co/facebook/nllb-200-distilled-1.3B

| **Bible - The New Testament** | |
|---|---|
| **1.** The book of the generation of Jesus Christ, the son of David, the son of Abraham. <br> **2.** And being warned of God in a dream that they should not return to Herod, they departed into their own country another way. | **1.** इब्राहीमः सन्तानो दायूद् तस्य सन्तानो यीशुख्रीष्टस्तस्य पूर्व्वपुरुषवंशश्रेणी । <br> **2.** पश्चाद् हेरोद् राजस्य समीपं पुनरपि गन्तुं स्वप्न ईश्वरेण निषिद्धाः सन्तो ऽन्येन पथा ते निजदेशं प्रति प्रतस्थिरे । |
| **Mann Ki Baat** | |
| **1.** My dear countrymen, namaskar! <br> **2.** A while ago, I had a chance to have an indirect dialogue, with young friends from Karnataka. | **1.** मम प्रियाः देश-वासिनः ! नमस्कारः । <br> **2.** कतिपय-दिनेभ्यः पूर्वम् अहं कर्णाटकस्य बालमित्रैः सह परोक्ष-संवादस्य अवसरं लब्धवान् । |
| **Gita Sopanam** | |
| **1.** Father comes home from office. <br> **2.** Paternal grandfather came home from library. | **1.** पिता कार्यालयात् गृहम् आगच्छति । <br> **2.** पितामहः ग्रन्थालयात् आगतवान् । |
| **Spoken Tutorials** | |
| **1.** Let's start today's tutorial with this image. <br> **2.** Today I will work with this image only to use it as an example. | **1.** अनेन चित्रेण वयम् अद्यतन पाठस्य आरम्भं कुर्मः । <br> **2.** अद्य, केवलम् उदाहरणत्वेन उपयोक्तुम् अनेन चित्रेण सह कार्य करोमि । |
| **NIOS** | |
| **1.** Narrate the context of the Ramayana. <br> **2.** understand the qualities of Rama. | **1.** रामायणस्य कथां ज्ञापयितुम्य । <br> **2.** रामस्य धार्मिकगुणांश्रावगन्तुम् । |

Table 5: Samples from different subsets of the Sāmayik.

## 4. Conclusion

We release a novel dataset, Sāmayik comprising of around 53,000 sentences for English-Sanskrit translation. Unlike existing datasets, Sāmayik emphasizes on contemporary prose writing and is curated from five diverse domains including instruction material, radio-podcast, *etc.* We also release a set of strong baselines built on four multilingual pre-trained models. We empirically demonstrate that models trained using our dataset achieve better performance than models trained on existing datasets, as well as pre-trained models incorporated with a Sanskrit corpus.

### Acknowledgements

We would like to thank following translators and reviewers towards development of the corpus.

1. Dr. Dinesh Joshi
2. Dr. Vasudev Aital
3. Shruti Sharma
4. Ashwini PN
5. Atmarama Bhat K

Ayush Maheshwari was supported by a fellowship from the Ekal Foundation during his Ph.D. at IIT Bombay. Ashim Gupta is supported by the Bloomberg Data Science Ph.D. Fellowship. This work is a result of the funding by the IKS Division of the Ministry of Education (MoE), Government of India to the IKS Projects (AICTE/IKS/RFP1/2021-22/05). Ganesh Ramakrishnan is also grateful to the National Language Translation Mission (NLTM): Bhashini project by Government of India and IIT Bombay Institute Chair Professorship for their support and sponsorship.